\pdfoutput=1
\documentclass[11pt]{article}

\usepackage[final]{acl}

\makeatletter
\AddToHook{env/tabular/before}{%
  \let\orig@startpbox@action\@startpbox@action
  \let\@startpbox@action\@startpbox
}
\AddToHook{env/tabular/after}{%
  \let\@startpbox@action\orig@startpbox@action
}
\makeatother
\usepackage{times}
\usepackage{latexsym}

\usepackage[T1]{fontenc}
\usepackage[utf8]{inputenc}

\usepackage{inconsolata}

\usepackage{graphicx}
\usepackage{multirow} 
\usepackage{amsmath}
\usepackage{array}
\usepackage{tabularx}
\usepackage{booktabs}
\usepackage{subcaption}
\usepackage{tcolorbox}
\usepackage[table]{xcolor}
\usepackage{pifont}
\usepackage[dvipsnames]{xcolor}
\usepackage{enumitem}
\usepackage{cellspace} 
\usepackage{makecell}
\usepackage{comment}
\usepackage{soul}
\usepackage{needspace}

\setcellgapes{1pt}
\makegapedcells
\title{\textsc{FicSim}: A Dataset for Multi-Faceted Semantic Similarity in Long-Form Fiction}

\author{
  Natasha Johnson\textsuperscript{1} \quad Amanda Bertsch\textsuperscript{1} \quad Maria-Emil Deal\textsuperscript{2} \quad Emma Strubell\textsuperscript{1} \\
  \textsuperscript{1} Language Technologies Institute, Carnegie Mellon University
  \\ \textsuperscript{2} School of Library and Information Studies, University of Oklahoma\\
\texttt{nmj@alumni.stanford.edu} \quad \texttt{abertsch@cs.cmu.edu} %\quad \texttt{strubell@cmu.edu}
}

\newcommand{\cmark}{{\protect\color{ForestGreen}{\ding{51}}}}
\newcommand{\xmark}{{\protect\color{BrickRed}{\ding{55}}}}

\definecolor{lightyellow}{rgb}{1,1,0.6}
\definecolor{bananamania}{rgb}{0.98, 0.91, 0.71}
\definecolor{blond}{rgb}{0.98, 0.94, 0.75}
\definecolor{champagne}{rgb}{0.97, 0.91, 0.81}
\definecolor{bisque}{rgb}{1.0, 0.89, 0.77}
\definecolor{eggshell}{rgb}{0.94, 0.92, 0.84}
\definecolor{lemonchiffon}{rgb}{1.0, 0.98, 0.8}
\definecolor{mistyrose}{rgb}{1.0, 0.89, 0.88}
\definecolor{lavendermist}{rgb}{0.9, 0.9, 0.98}
\definecolor{papayawhip}{rgb}{1.0, 0.94, 0.84}
\sethlcolor{papayawhip}

\begin{document}
\maketitle
\begin{abstract}
As language models become capable of processing increasingly long and complex texts, there has been growing interest in their application within computational literary studies. However, evaluating the usefulness of these models for such tasks remains challenging due to the cost of fine-grained annotation for long-form texts and the data contamination concerns inherent in using public-domain literature.  
Current embedding similarity datasets are not suitable for evaluating literary-domain tasks because of a focus on coarse-grained similarity and primarily on very short text. 
We assemble and release \textsc{FicSim}, a dataset of long-form, recently written fiction, including scores along 12 axes of similarity informed by author-produced metadata and validated by digital humanities scholars. We evaluate a suite of embedding models on this task, demonstrating a tendency across models to focus on surface-level features over semantic categories that would be useful for computational literary studies tasks.
Throughout our data-collection process, we prioritize author agency and rely on continual, informed author consent.\footnote{There are slight discrepancies between the results reported in this version of the paper and the results published in \textit{Findings of the ACL: EMNLP}. These discrepancies are due to an error made when embedding 2 of the fanfics with the Voyage-3-large and m2-BERT-32k  models. The results reported in this paper are the corrected results.}\footnote{Dataset can be accessed at \url{https://huggingface.co/datasets/ficsim/ficsim}. Additional documentation can be found at \url{https://github.com/natashamariejohnson330/FicSim}} 
\end{abstract}

\section{Introduction}
\begin{figure}[h]
    \centering
    \includegraphics[width=0.95\linewidth]{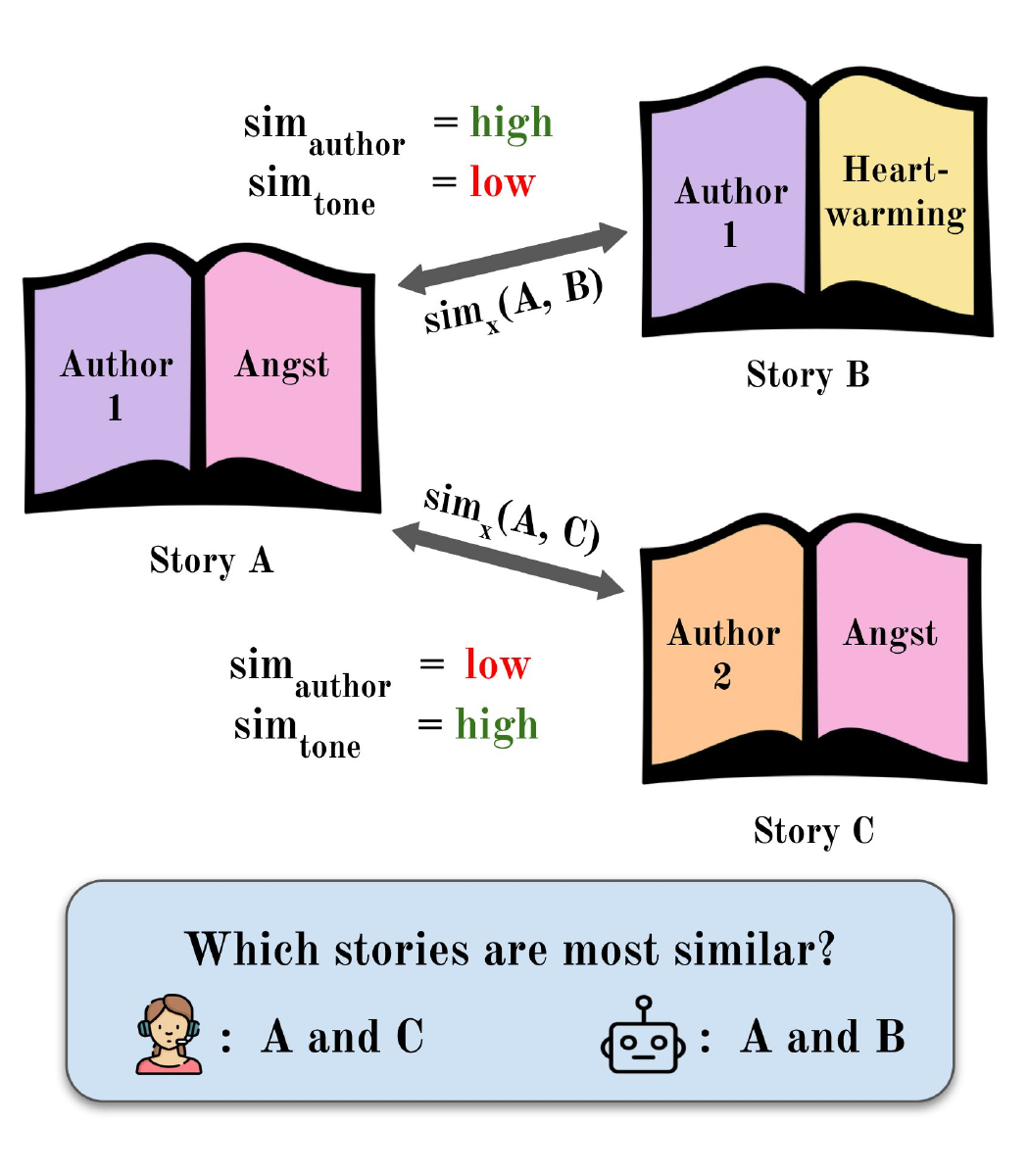}
    \caption{Similarity between literary texts can be defined along many dimensions. Computational literary studies scholars generally seek to measure specific, semantic types of similarity such as similarity in tone, but embedding models over-index on more obvious features such as the author's writing style.}
    \label{fig:enter-label}
    
\end{figure}
The last few years have been a time of immense progress in long-context processing in NLP. Several language models now support context lengths in excess of a million tokens. Embedding models for 32k context inputs abound. While challenges remain in long context modeling, successive approaches have made strong progress on benchmarks and have found applications downstream \cite{kapoor2024promisespitfallsartificialintelligence, godbole2024leveraginglongcontextlargelanguage, nie2024surveylargelanguagemodels}.

In parallel, there is interest in applying NLP methods within digital humanities (DH), particularly within computational literary studies. Many DH scholars have incorporated NLP methods such as topic modeling, sentiment analysis, and semantic textual similarity (STS) tasks such as clustering and measuring similarity into  their research \citep{kleymann-2022, algee-hewitt-2023}.

\begin{table}
\setlength{\tabcolsep}{2pt}
\renewcommand{\arraystretch}{.8}
\scriptsize
  \centering
  \begin{tabular}{p{2cm}|ccccc}
    \toprule
    \textbf{Source} & \textbf{Long} & \shortstack{\textbf{Publicly} \\ \textbf{available}} & \shortstack{\textbf{Not publicly} \\ \textbf{analyzed}} & \shortstack{\textbf{Author} \\ \textbf{labeled}} & \shortstack{\textbf{Multiple axes} \\ \textbf{of similarity}} \\
        \midrule
    Recent novels & \cmark & \xmark & \cmark  & \xmark & \xmark \\
    Project Gutenberg & \cmark & \cmark & \xmark & \xmark & \cmark \\
    MTEB STS tasks & \xmark & \cmark  & \cmark  & \xmark & \xmark \\
    AO3 Fanfiction & \cmark & \cmark & \cmark & \cmark & \cmark  \\
    
    \bottomrule
  \end{tabular}
  \caption{\label{tab:desiderata}
    Comparison of data sources for semantic textual similarity tasks. Fanfiction represents our approach. 
  }
\end{table}

Yet evaluations of these approaches have been limited, particularly with regards to STS tasks. Many NLP models which have been (or have the potential to be) applied in DH research are not evaluated on literary applications. 
When they are, it is often on digital texts made accessible through public repositories such as Project Gutenberg \citep{kohlmeyer-2021, kryscinski-2019, underwood-2018, bamman-2024, kocisky-etal-2018-narrativeqa, xu-etal-2022-fantastic}. However, these texts---alongside related data and analyses from other commonly-scraped sites like Wikipedia---are included in the pretraining data of most models \citep{elazar-2024}, which could cause direct or indirect contamination and skew evaluation results \citep{palavalli-2024, zhang-2024}.  

Furthermore, while existing literary datasets evaluate model suitability for tasks such as summarization \cite{kryscinski-2019}, question answering \cite{kocisky-etal-2018-narrativeqa,xu-etal-2022-fantastic}, and identifying literary co-reference \cite{bamman-etal-2020-annotated}, DH scholars have expressed the need for embedding methods that capture semantic textual similarity within novel-length texts along several axes such as plot, tone, and setting \cite{sobchuk-2024}.

In response to this gap, we present \textsc{FicSim}, an evaluation dataset for fine-grained semantic textual similarity (STS), constructed of long-form human-written narratives that are unlikely to appear in pretraining data,  are accompanied by author-labeled metadata, and are included in this dataset \textit{with \mbox{author} consent}.
We describe our processes for selecting text not included in CommonCrawl scrapes, for obtaining and maintaining author consent for the use of their works, and for constructing pairwise similarity measurements corresponding to 12 different facets of fictional texts, in consultation with both literary scholars and authors (\S \ref{sec:dataset-building}). We then describe the resulting dataset and its use (\S \ref{sec:dataset-using}). We evaluate existing approaches on this multi-faceted STS task (\S \ref{sec:methods}). Models struggle to capture salient characteristics of long-form texts, only weakly disambiguate between categories, and over-index on surface features of the text (\S \ref{sec:results}). We conclude by discussing the relevance of our results for both literary studies scholars and NLP researchers  (\S\ref{sec:conclusion}). We hope that \textsc{FicSim} enables more focus on narrowing the gap between models' general capabilities and their applicability to literary domain tasks.

\section{Sourcing Data}
\label{sec:dataset-sourcing}
\subsection{Desiderata}
 To effectively measure similarity in long-form fiction, we need a corpus of stories with several characteristics, summarized in Table~\ref{tab:desiderata}: The stories must be (1) \textit{long}, coherent narratives  (2) \textit{publicly available} for evaluation (3) \textit{not publicly analyzed online}, in order to prevent potential contamination (4) \textit{well-annotated}, ideally by an expert, capturing \textit{multiple axes of similarity}, beyond superficial similarities that are easy to detect but limited in the value for literary scholars.

Recently published novels offer one compelling solution \cite{karpinska-etal-2024-one, duarte-2024} but limit public release of the dataset.\footnote{Copyright laws do permit physical copies of books to be purchased, scanned, and shared with certain provisions. However, digitizing physical texts can be a labor and cost intensive.} Furthermore, the suitability of such novels for evaluation purposes decays over time, as summaries and analyses of the texts become increasingly likely to have been incorporated into model training. Public domain literature, such as the texts made available through Project Gutenberg, satisfies the length and public availability requirements, but is deeply present both in pretraining corpora and in public culture---with many analyses online of the themes, plot, and character arcs of each story, it is unclear whether a model identifying these characteristics is doing so through memorization \cite{palavalli-2024}. %

\subsection{Our Approach} In response, we turn towards \textit{fanfiction}--fictional texts inspired by existing media, often sharing the characters or setting of the source work. Many fanfiction texts are complex long-form narratives, reflecting both the fandom subculture and major cultural movements of the time. In recent decades, fan studies has become an active subfield within literary, media, and cultural studies. 

\begin{figure*}
    \centering
    \includegraphics[width=1\linewidth]{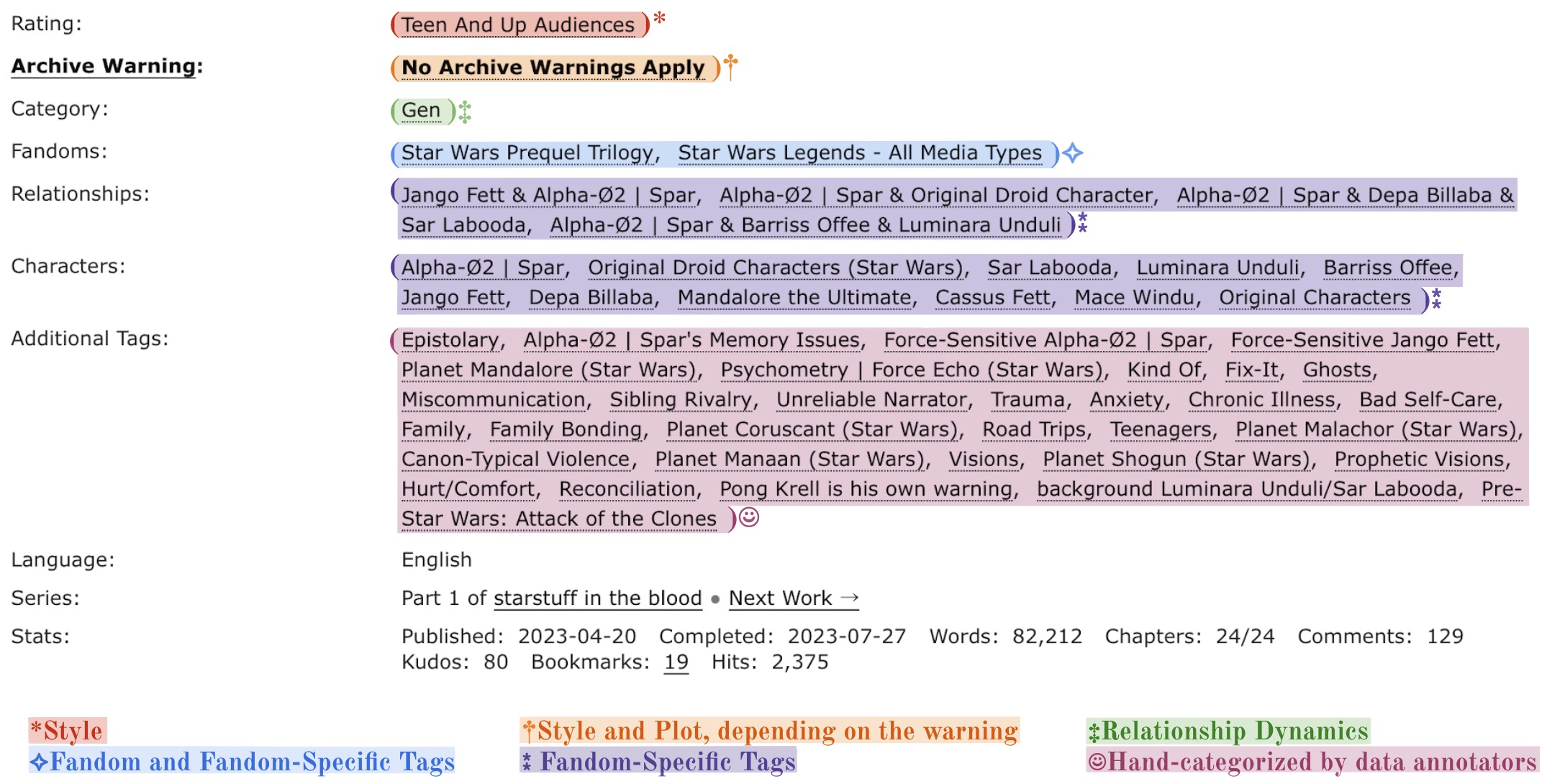}
    \caption{Example of fic tagging and metadata on AO3. Colored annotations mark data that inform similarity scores.
    }
    \label{fig:detailed_tagging}
\end{figure*}

Significantly for our purposes, popular fanfiction websites allow users to assign tags to their fanfiction (e.g. Figure~\ref{fig:detailed_tagging}) to help with fanfic discoverability and categorization. Tags range from purely descriptive to analytical to conversational and identify important elements of a fanfic from the author's perspective. They are intended to help readers find the content that is interesting to them among millions of stories. We use such tags as the basis for our gold-standard similarity scores.

\subsection{Story Selection}

\paragraph{Archive of Our Own.} We source our fanfiction from \href{https://archiveofourown.org/}{Archive of Our Own} (AO3), a digital repository hosting over 15M works. We selected this venue for two main reasons: First, AO3 has made significant efforts to discourage web scraping, including blocking Common Crawl scraping in 2022 \citep{organization-for-transformative-works-2023} and implementing aggressive rate limiting policies.\footnote{While these do not \textit{guarantee} that no fanfics posted after this date are used in pretraining corpora, these restrictions, along with AO3's lack of an official API, make these texts relatively unlikely to appear in pretraining web scrapes.}  Second, due to the site's construction and norms, many AO3 stories feature highly detailed tagging (e.g. Figure~\ref{fig:detailed_tagging}), which we leverage to compute similarity along various axes.

\paragraph{Requirements for stories.} We only consider stories that are written in English, exceed 10,000 words, and feature detailed tagging. Furthermore, because AO3 restricted web crawling in December 2022, we only consider stories that were started and completed after this date. %We also only consider completed works. We restrict to stories written in English that exceed 10,000 words, and select stories that have detailed tagging that is not specific to only a particular fandom. 
Focusing on texts over 10,000 words ensures that the stories in our dataset are similar in length to fictional texts commonly studied by literary scholars, including short stories, novelettes, novellas, and novels \cite{gioia-2006, harmon-2003}. We do not screen for or remove explicit content, although the majority of the stories in the dataset are not marked as explicit.\footnote{Explicit texts (which may be tagged as such because of violent or sexual content) are an active area of research in media studies (e.g. \citet{erotic-fanfic-activism-thesis, explicitfanficthesis}).} Additionally, when identifying stories that we hoped to include in our dataset, we looked for a variety of tropes, settings, and writing styles.

\subsection{Author Consent}
Many artistic communities have strong negative feelings towards the machine learning community. Writers have described the AI community as exhibiting ``complete lack of respect'' for artistic work \citep{Gero_2025}. 
One survey found that 96\% of authors were against the use of their work for AI training without their explicit consent \citep{authorsguild2023ai}. Fanfiction authors are no exception: fan communities have expressed dismay over learning that some fanfiction has been included in CommonCrawl datasets, and Archive of Our Own lawyers went before the U.S. Copyright Office to argue that fan authors should be able to opt-out their work from model pretraining corpora \cite{ao3ai2023statement}.

Therefore, although Archive of Our Own explicitly permits metadata collection done by academic researchers \citep{organization-for-transformative-works-2023}, we additionally received each author’s permission to include their work(s) in our dataset using an IRB-approved process. We reached out to each author individually through AO3 to explain our project and share guarantees about author privacy, story usage, and withdrawal of consent.  Then, we invited authors to sign a consent form and provided mechanisms to ask questions or withdraw consent at any time.  
Furthermore, in consultation with several members of fan communities, we decided to commit to not using \textsc{FicSim} to train models; to access the dataset, we will require other researchers to agree to the same terms of usage.
For more information on the consent process and the full text of the outreach documents, see Appendix~\ref{appendix:author-consent-forms}.

\subsection{Fanfiction versus Traditional Literature}
Over the past decade in particular, the line between fanfiction and published literature has been blurred in terms of both story content and writing style. Many stories originally written as fanfiction have been published as standalone books (with minimal edits, such as changing the characters’ names) and have seen commercial success \cite{arzbaecher-2023}. Many tropes created or popularized by fan communities have been adopted outside of fan spaces by authors and consumers of contemporary genre fiction \cite{majnaric, jerasa-2021}. Furthermore, when performing stylistic comparison between fanfiction stories and their inspiration texts (e.g. Harry Potter novels), researchers have found that fanfiction stories do not stylistically diverge from their source material in statistically significant ways \cite{jacobsen-2024}. Thus, we believe fanfiction texts are suitable for evaluating model performance on literature, particularly contemporary fiction. 

\subsection{Use of Fanfiction Tags}

We derive gold-standard annotations for document similarity from the tags each author has ascribed to their story, which may be either canonical \footnote{That is, standardized tags that are internally linked to synonymous and related tags by AO3's tag-wrangling team.} or user-authored.  

Our use of user-generated tags as the basis for our similarity calculations was informed by the work of \citet{lyons-2008}, who explored using user-generated tags to expand access to library resources. They discuss how user-based tagging offers a form of natural language keyword categorization that can help capture particular narrative features which are not standard subject headings for narrative works. In this way, user-generated tags do a better job of helping guests find materials that relate to their interests.
This is essentially the purpose that tags serve on AO3---its tagging system, handling of user-generated tags, and user norms have allowed readers to find the content that is interesting to them among millions of other stories.

Though AO3 tags might seem like a limited source from which to construct similarity scores,
within DH work, genre labels are often used as ground-truth labels for evaluating literary embedding, topic modeling, and clustering methods \citep{sobchuk-2024, schoch-2017, Allison}. Within fan spaces, fanfiction tags serve many of the same purposes and are held in the same common understanding \citep{hellekson-2014} as genre labels. In fact, many tags created or popularized by fan communities such as ``Enemies to Lovers'' and ``Slow burn'' have notably been adopted outside of fan spaces by authors and consumers of contemporary genre fiction \citep{majnaric}. These tags are commonly used in advertisements, recommendations, and reviews of books  
\citep{jerasa-2021}. Goodreads has even added "Enemies to Lovers" as a book genre on their site \citep{goodreads}.

 Thus, our process calculates story similarity according to a well-established framework of what is important in a story, as developed and refined by fanfiction authors and readers, and as adopted by many traditional authors, publishers, and readers.

\section{Constructing \textsc{FicSim}}
\label{sec:dataset-building}
 
\subsection{Tag Categorization} 

We place tags into 12 different categories, corresponding with various types of similarity we hope to measure. Some categories, such as ``Plot,'' ``Theme,'' and ``Time'' capture general qualities of fictional literature and align with projects that discuss narrative similarity on the basis of actions, subjects, themes, and temporal setting \citep{algee-hewitt-2023, kleymann-2022, sobchuk-2024, piper-2022}. Others, such as ``Fanfiction Tone and Content Tags'' capture fanfiction-specific qualities and serve to identify whether models can identify similarity based on genre-specific conventions. To determine the categories, we relied on the aforementioned DH scholarship, as well as the annotators' analysis of the tag set. When applicable, we place tags into multiple categories. Table~\ref{tab:tag_categories} describes the categories with tag examples from each. 

\begin{figure*}[t]
\centering
\begin{minipage}{0.95\textwidth}
  \small
  \centering
  \begin{tabular}{l p{0.45\linewidth} l}

    \toprule
    \textbf{Category} & \textbf{Description} & \textbf{Example Tags} \\
    \midrule 
    \textbf{Plot} & Narrative actions and concrete subjects; describes what happens in a story and what is \textit{in} a story. & \makecell[l]{Letters \\ Blood and Torture \\ Artificial Intelligence} \\ \hline
    \textbf{Character States} & The emotions, attributes, roles, and physical characteristics of characters in the text.& \makecell[l]{Trans Woman \\ Character Is Bad at Feelings \\ Dissociative Identity Disorder} \\ \hline
    \textbf{\makecell[l]{Relationship \\ Dynamics} } &  Key characteristics of both platonic and romantic relationships in the text. & \makecell[l]{Established Relationship \\ Possibly slowest ever burn     \\ F/F} \\ \hline
    \textbf{Theme} & Abstract ideas explored throughout the story.& \makecell[l]{Racism \\ Found Family  \\ What Is The Impact Of A Mother} \\ \hline
    \textbf{Time} & Temporal setting.& \makecell[l]{Alternate Universe - 19th Century \\ Modern Retelling \\ Post-Apocalypse } \\ \hline
    \textbf{Style} & Features of the writing style or narrative technique.  & \makecell[l]{POV Third Person Omniscient \\ Epistolary  \\ Dialogue Heavy } \\ \hline
    \textbf{\makecell[l]{Fanfiction Tone \\ and Content Tags}} & Fan-community language for the type of story; often relates to both tone and plot (e.g. ``fluff'' generally involves lighthearted and domestic scenes). & \makecell[l]{Angst \\ Fluff and Hurt/Comfort \\ Tooth-Rotting Fluff } \\ \hline
    \textbf{Fandom-specific} & Tags that involve settings, characters, or events from a canon text, and thus reveal information about the fandom the story belongs to. & \makecell[l]{Phantom of the Opera AU  \\ Yule Ball (Harry Potter) \\ Capitano/Mavuika } \\ \hline\textbf{\makecell[l]{Overall \\(Fandom-Agnostic)}} & An aggregate grouping of tag categories 1-7; captures similarity between all tags that do not reveal the fandom or author identity. & \makecell[l]{[any of the above tags, excluding \\fandom-specific]}\\
    \hline 
    \textbf{\makecell[l]{Overall \\(Fandom-Specific)}} & An aggregate grouping of tag categories 1-8; captures similarity between all tags & [any of the above tags] \\  \hline
    \textbf{Fandom} & Captures whether two texts are inspired by the same piece(s) of media (e.g. books, movies, television shows). 
    &  \makecell[l]{Genshin Impact \\ Grey's Anatomy\\Star Wars}\\  \hline
    \textbf{Author} & Captures whether two texts are written by the same fan author. & [author IDs]\\
    \bottomrule
  \end{tabular}
  \captionof{table}{\label{tab:tag_categories}
    Types of similarity and example tags that align with each category. Some of the above tags were placed into multiple categories within our dataset, but they nonetheless serve as useful references for the particular category they represent above.
  }
\end{minipage}
\end{figure*}

\paragraph{Tag cleaning.} In order to increase tag interpretability and allow for clearer tag comparison, we clean and standardize the tags prior to constructing similarity scores. The cleaning process involves removing unnecessary punctuation, standardizing capitalization, and correcting obvious spelling errors. Our tag standardization involves replacing tags with their canonical counterparts and rephrasing or restructuring tags to increase their semantic interpretability.  

\Needspace{2\baselineskip}

\paragraph{Fandom-agnostic tags.} In order to allow our dataset to evaluate model suitability for applications outside of fan studies, we want all tag categories except Fandom, Fandom Specific, and Overall (fandom-specific) to contain only fandom-agnostic terms which could be used to compare story similarity across fandoms. For the nearly 50\% of the tags that contained fandom-specific references, we create a fandom-agnostic version of the tag by removing or replacing fandom-specific references (e.g. ``Protective Cristina Yang'' becomes ``protective character''). We then place this fandom-agnostic tag into the appropriate fandom-agnostic category(/ies) while also placing the original tag into the Fandom and Fandom Specific category. 

\paragraph{Annotation process.} Our tag categorization and rewriting was performed by two authors who are experts in fan studies, one with a background in computational literary studies and the other with a background in library and information science. Following the qualitative methods tradition of interpretative analytical process ~\cite{lincoln2011paradigmatic,strauss1990basics}, the annotating authors arrived at annotations through a series of consensus-building discussions.  In a few instances where tags contained fandom-specific terminology for a particular fandom that neither annotator had specific expertise in, the annotators reached out to fanfiction authors within that fandom to confirm their interpretation of the tag prior to categorization. In these cases, the authors were acquaintances that each annotator knew through their own engagement with fan communities.

\subsection{Similarity Score Calculation}

After tags are cleaned and categorized, we calculate category-specific similarity scores between stories. We embed each tag using Gemini Embedding \cite{lee-2025}
. Then, given category-specific tag lists from two fanfics, we calculate the similarity between tag lists $A$ and $B$
as the average of the pairwise tag cosine similarities: \[
\text{sim}(A, B) =  \operatorname{avg} \text{sim}_{a \in A, b \in B}(a, b)
\]

\paragraph{Gemini embedding.} We selected Gemini Embedding as our embedding model for the following reasons: (1) It has the overall highest score on the MTEB leaderboard \cite{muennighoff-etal-2023-mteb}; (2) Among the best-performing models on the MTEB English-language STS tasks\footnote{As of April 2025.}, Gemini Embedding is the highest ranking model that is not from the same model family (or built upon the same model family) as the models we evaluate below. Furthermore, because Gemini Embedding has an input limit of 2048 tokens, selecting it as our tag embedding model does not then preclude us from including it in our evaluation section. 

\paragraph{Gold score validation.} We validate our tag handling and similarity score construction in two steps. Prior to tag cleaning, one author annotated a set 330 story triplets (thirty 3-way story comparisons in each of the 11 non-author categories), identifying whether the first story in a triplet was more similar to the second or third story from that triplet in a given category.\footnote{Story triplets were drawn from a randomly-generated set of 100 story triplets in each category. The author identified the first 30 non-ambiguous comparisons in each category, skipping  triplets where the author thought an argument could be made for either ranking.} Tag cleaning, standardization, and embedding processes were then adjusted to align the resulting gold similarity scores with the author's annotations. Then, to further validate the scores, two other authors each annotated identical sets of 220 story triplets (20 stories x 11 categories) on the same task. Since the two annotators were not both experts in fan studies or literature, they were allowed to skip comparisons for which they did not think they could identify the more similar story. Of the 158 triplets that neither annotator skipped we measured annotator agreement of 82\% (Cohen's $\kappa$ = 0.65). Our gold truth labels were then evaluated against the 129 triplets for which both authors provided the same rating, demonstrating 80\% alignment with the annotations.

\subsection{\textsc{FicSim} Statistics}

\label{sec:dataset-using}
Our final dataset includes 90 stories\footnote{Literary datasets with expert annotations are often comparable in magnitude \cite{sims-2019, bamman-etal-2020-annotated}.} and gold similarity scores along 12 axes, for a total of 33,790 pairwise comparisons. The stories range from 10,001 to 488,772 words and span 46 fan communities (fandoms). The cleaned tagset has 9448 total tags (2133 unique) across 12 categories.  
For more detailed dataset statistics and license information, see Appendices~\ref{appendix:datastats} and ~\ref{appendix:license}.

\subsection{Similarity scores}
For every story pair, \textsc{FicSim} presents up to 12 similarity scores, described in detail in Table~\ref{tab:tag_categories}. We divide these scores into three evaluation groups:

\paragraph{Fine-grained similarity:} Plot,  character states, relationship dynamics, theme, time, style, fanfiction tone \& content tags.

\paragraph{Broader notions of similarity:} Overall (fandom-agnostic and fandom-specific).  The overall (fandom-specific) category captures a mix of both superficial and more integrated narrative elements, while the overall (fandom-agnostic) category primarily targets the latter.

\paragraph{Superficial similarity:} Fandom-specific tags, fandom, and author. Because these are always provided (even when a story is not otherwise tagged well) and generally obvious to deduce from the text, we do not consider these axes of similarity to evaluate embedding quality, only to measure how embeddings capture these in contrast to the more fine-grained features above.

\section{Evaluation}
For each model, we compute cosine similarity between embeddings for all story pairs. Following \citet{muennighoff-etal-2023-mteb}, we measure Spearman's $\rho$ between the model-induced ranking of story similarities and our tagset-derived gold ranking.  
We report  $\rho$ out of 100 instead of 1.0 for readability, and highlight values that are significant $(p < 0.05)$.

\label{sec:methods}

\begin{table*}
  \renewcommand{\arraystretch}{0.2}
  \setlength{\fboxsep}{0pt}
  \small
  \centering
  \addtolength{\tabcolsep}{-0.2em}
  \begin{tabular}{lrrrrrrr|r|rrrr}
    \toprule
    \textbf{Model} &  \scriptsize \textbf{Plot} & \scriptsize \textbf{\shortstack{Char-\\acter \\ States}} & \scriptsize \textbf{\shortstack{Rela-\\tionship \\ Dynamics}} & \scriptsize \textbf{Theme} & \scriptsize \textbf{Time}  & \scriptsize \textbf{Style} & \scriptsize \textbf{\shortstack{Fanfiction\\Tone \& \\ Content}} & \scriptsize \textbf{\shortstack{All\\ (Fandom\\ Agnostic)}} & \scriptsize \textbf{\shortstack{All\\ (Fandom-\\Specific)}} & \scriptsize \textbf{\shortstack{Fandom \\Specific}}  & \scriptsize \textbf{Fandom}  & \scriptsize \textbf{Author} \\
    \hline
   \scriptsize Linq-Embed & \cellcolor{blue} \hl{\textbf{18.65}} & \hl{8.63} & \hl{5.52} & 0.14 & \hl{28.85} & 3.33 & \hl{9.95} & \hl{16.59} & \hl{34.61} & \hl{34.99} & \hl{30.47} & \hl{\textbf{40.91}} \\ 
\qquad+ SW & \hl{10.17} & \hl{5.69} & \hl{9.12} & \hl{6.87} & \hl{21.51} & \hl{7.03} & \hl{6.09} & \hl{7.37} & \hl{15.17} & \hl{18.02} & \hl{29.25} & \hl{40.73} \\ 
\scriptsize GTE-Qwen2 & \hl{15.84} & \hl{4.36} & \hl{9.62} & \hl{\textbf{13.92}} & \hl{31.23} & \hl{13.43} & \hl{\textbf{20.87}} & \hl{24.38} & \hl{40.50} & \hl{45.48} & \hl{45.87} & \hl{40.82} \\ 
\qquad+ SW & \hl{13.00} & -1.01 & \hl{10.74} & \hl{7.52} & \hl{24.84} & \hl{12.96} & 1.74 & \hl{11.02} & \hl{20.49} & \hl{20.65} & \hl{26.73} & \hl{40.73} \\ 
\scriptsize SFR-Embed & \hl{17.06} & 0.98 & \hl{5.83} & 4.96 & \hl{24.34} & -5.06 & \hl{8.45} & \hl{18.07} & \hl{40.65} & \hl{36.62} & \hl{34.35} & \hl{38.76} \\ 
\qquad+ SW & \hl{10.42} & \hl{7.14} & \hl{10.73} & \hl{7.98} & \hl{22.66} & \hl{9.28} & \hl{8.05} & \hl{10.47} & \hl{17.88} & \hl{19.29} & \hl{31.49} & \hl{40.64} \\ 
\scriptsize GTE-ModernBERT & \hl{13.49} & \hl{\textbf{11.87}} & \hl{15.39} & 0.17 & -0.30 & \hl{8.20} & \hl{13.28} & \hl{16.77} & \hl{42.70} & \hl{48.50} & \hl{\textbf{50.08}} & \hl{36.33} \\ 
\qquad+ SW & \hl{18.07} & \hl{4.68} & \hl{\textbf{23.82}} & \hl{6.71} & \hl{\textbf{40.24}} & \hl{16.43} & \hl{5.96} & \hl{22.13} & \hl{41.37} & \hl{42.03} & \hl{46.57} & \hl{37.60} \\ 
\scriptsize m2-BERT-32k & \hl{5.09} & \hl{7.41} & \hl{12.87} & 4.20 & -2.21 & -3.38 & \hl{11.66} & \hl{7.80} & \hl{15.58} & \hl{14.19} & \hl{11.03} & \hl{19.59} \\ 
\qquad+ SW & \hl{8.96} & \hl{7.77} & \hl{13.54} & \hl{6.35} & 0.90 & 5.00 & \hl{19.65} & \hl{12.81} & \hl{20.68} & \hl{20.02} & \hl{22.87} & \hl{20.18} \\ 
\scriptsize Voyage-3-large & \hl{13.40} & \hl{5.72} & \hl{9.86} & \hl{7.11} & \hl{29.95} & \hl{\textbf{22.51}} & \hl{15.40} & \hl{\textbf{25.08}} & \hl{\textbf{50.95}} & \hl{45.92} & \hl{37.09} & \hl{40.62} \\ 
\qquad+ SW & \hl{14.10} & \hl{4.35} & \hl{8.07} & \hl{6.90} & \hl{28.38} & \hl{20.63} & \hl{12.67} & \hl{24.41} & \hl{50.20} & \hl{47.67} & \hl{39.55} & \hl{40.59} \\ 
\scriptsize Claude+SW & -1.19 & -0.74 & \hl{5.94} & \hl{11.64} & \hl{27.54} & \hl{17.05} & -0.43 & \hl{8.99} & \hl{39.27} & \hl{\textbf{49.41}} & \hl{44.95} & \hl{40.19} \\
    \bottomrule
  \end{tabular}
  \caption{\label{tab:overall-embedding}
    Spearman correlation of embedding cosine similarity to our tagset similarity measures, for several representative open-source and API-based embedding models. All models struggle at category similarity and overindex on authorial style. +SW denotes the use of a sliding window; Statistical significance \hl{highlighted}.
  }
\end{table*}

\paragraph{Models.} We select the 3 best-performing open-weights models on the Hugging Face \cite{wolf2020huggingfacestransformersstateoftheartnatural} MTEB leaderboard with 32k context lengths: Linq-Embed-Mistral \cite{choi2024linqembedmistraltechnicalreport}, GTE-Qwen2-7B-instruct \cite{li2023towards}, and SFR-Embed-Mistral \cite{SFRAIResearch2024}.  
These are all 7B models; however, computational literary studies scholars often have access to limited computational resources. Thus, we consider a much smaller model-- GTE-ModernBERT-base \cite{zhang2024mgte}-- and two API-based solutions: m2-BERT-80M-32k-retrieval \cite{fu2023monarch} through the TogetherAI API and using Voyage-3-large \cite{voyage3large2025} through the Voyage.ai API.  Finally, we consider whether large language models could perform this task.  
We use Claude-3.7-Sonnet \cite{anthropic2025claude37} to summarize each story, then embed the much shorter summary documents with Voyage.\footnote{Anthropic does not have its own embedding model; we use Voyage on the Claude outputs because this is the embedding model Anthropic recommends in \href{https://docs.anthropic.com/en/docs/build-with-claude/embeddings}{their documentation}.}
We follow each model's default strategy for constructing embeddings; for additional details on models and pooling methods, see Appendix~\ref{appendix:models}.

\paragraph{Prompt.} Linq-Embed, SFR-Embedding, and both GTE models support providing an instruction in a special format at the beginning of a text to be embedded. We experiment with using this to produce category-specific embeddings, by providing instructions to focus on each type of similarity in turn. 
We provide the prompt (see Appendix~\ref{appendix:models}) to each embedding model at the start of the document.\footnote{For the other models that do not support a custom instruction, we simply prepend the prompt without special formatting.} For Claude, we use a modified form of category-specific prompt for summarizing and omit the prompt for the embedding stage.

\paragraph{Sliding window.} There are stories in \textsc{FicSim} that exceed the context length of every model evaluated. When stories exceed the maximum context length, we consider two options. As a baseline, we naively truncate each story to the maximum length.  
We then consider a sliding-window approach where we chunk the text to windows of the maximum context length and then pool embeddings across windows \cite{wang-etal-2019-multi}. 
For the sliding window approach, we overlap windows by 2048 characters (depending on the model and story, approximately 500 words).

\section{Results: Can existing models perform fine-grained literary STS?}
\label{sec:results}

We present results using a single general-purpose embedding from each model in Table~\ref{tab:overall-embedding} and results on category-specific embeddings for each model in Table~\ref{tab:category-specific}. All models struggle to perform fine-grained STS; unexpectedly, the larger models are not consistently better than the small embedding models. We discuss overall trends below. 

\paragraph{Truncation and sliding windows show marginal differences.} Surprisingly, naively truncating to the first segment of the story and taking the mean of sliding window embeddings are similarly successful across the board. Though the mean Spearman correlation for truncation is slightly higher than the mean sliding window correlation (11.75 vs 11.69), when comparing individual correlations for a given model and category, truncation beats sliding window exactly 50\% of the time. \footnote{Note that we do not compare between sliding window and truncation for the Claude results because only four stories exceed Claude's 200k context window, so there is not enough data to make a meaningful comparison between the methods.}   
Despite allowing a model to consider more of the text, sliding window embeddings may fail to increase model performance because models are not generally trained for this approach; because the pooling strategy needs to be adjusted when pooling across more data; or simply because some of the features that capture story similarity can be extracted from the start of the text alone. That being said, while sliding windows do not improve correlation scores in any of the fandom-agnostic categories, they do decrease correlations in the confounding categories (mean $\rho$ of 28.62 as opposed to 36.53), suggesting that their embeddings weigh surface-level features less heavily.

\begin{table}
  \small
  \centering
  \addtolength{\tabcolsep}{-0.22em}
  \begin{tabular}{lrrrr}
    \toprule
    \textbf{Gold score category} & \scriptsize\textbf{\shortstack{Overall \\(Fandom\\ Specific)}} & \scriptsize\textbf{\shortstack{Fandom\\Specific}}  & \scriptsize\textbf{Fandom}  & \scriptsize\textbf{Author} \\ \midrule
\scriptsize Plot & \hl{31.85} & \hl{7.56} & \hl{8.43} & \hl{12.14} \\
\scriptsize Character States & \hl{4.52} & 0.09 & \hl{15.10} & \hl{13.55} \\
\scriptsize Relationship Dynamics & \hl{36.11} & \hl{10.58} & \hl{17.25} & \hl{17.53} \\
\scriptsize Theme & \hl{10.30} & 1.95 & 3.98 & 5.10 \\
\scriptsize Time & 12.53 & 14.36 & \hl{\textbf{34.65}} & \hl{\textbf{21.28}}\\
\scriptsize Style & \hl{24.18} & \hl{8.90} & \hl{11.21} & \hl{13.05} \\
\scriptsize Fanfic Tone \&  Content & 00.00 & -3.61 & -1.05 & 5.19 \\
\scriptsize Overall (Fandom-Agnostic)& \hl{\textbf{48.01}} & \hl{\textbf{14.04}} & \hl{18.90} & \hl{19.20} \\
    \bottomrule
  \end{tabular}
  \caption{\label{tab:gold-confounder-correlation}
    Gold category-specific scores range in correlation with the overall (fandom-specific), fandom-specific, fandom, and author categories. The comparatively low scores in the latter three categories indicate that the gold scores weigh surface-level features far less heavily than our embedding models. Statistical significance \hl{highlighted}} 
\end{table}

\paragraph{Models overindex on surface features.} In the vast majority of cases, Spearman's $\rho$ is higher for the four ``confounder'' categories than any of the other similarity categories.  Notably, the author category (which is computed solely from exact match of author IDs) has the highest score more frequently than any other category. This indicates that embeddings are much more sensitive to author-specific stylistic factors than to the fine-grained semantic factors captured in the remaining categories. 

Some sensitivity to author and fandom is expected---some authors will focus on different types of stories, and some conventions or styles will be more common in one fandom than another. The overall (fandom-specific) category takes this into account by considering similarity based on both fandom-specific and non-fandom tags. However, 77\% of the models in Table \ref{tab:overall-embedding} score higher in the fandom-specific, fandom, or author category than in the overall (fandom-specific). This indicates that they are not only capturing the fingerprints of certain authors and fandoms, but that they are furthermore failing to capture other narrative elements. In contrast, Table~\ref{tab:gold-confounder-correlation} shows how the gold scores correlate with fandom and author-based categories. Many of the gold scores are positively correlated with the overall (fandom-specific) category, reflecting the fact that this category incorporates all other narrative features into its ranking. However, the majority are not as strongly correlated with the fandom-specific, fandom, and author scores, indicating that the information they capture extends beyond these categories. 

\begin{table*}
  \small
  \centering
  \addtolength{\tabcolsep}{00.00em}
  \begin{tabular}{lrrrrrrrr}
    \toprule
    \textbf{Model} & \textbf{Plot} & \textbf{\shortstack{Character \\ States}} & \textbf{\shortstack{Relationship \\ Dynamics}} & \textbf{Theme} & \textbf{Time}  & \textbf{Style} & \textbf{\shortstack{Tone \& \\ Content}} & \textbf{\shortstack{Overall \\(Fandom-Agnostic)}} \\
    \hline
    Linq-Embed & \hl{8.36} & \hl{6.00} & \hl{4.94} & 4.38 & \hl{24.96} & \hl{6.63} & \hl{9.90} & \hl{16.59} \\ 
\qquad+ SW & \hl{10.30} & \hl{5.90} & \hl{9.36} & \hl{7.03} & \hl{21.67} & \hl{7.30} & \hl{6.26} & \hl{7.37} \\ 
GTE-Qwen2 & \hl{15.44} & \hl{5.36} & \hl{9.43} & \hl{14.43} & \hl{33.19} & \hl{14.01} & \hl{\textbf{20.35}} & \hl{24.38} \\ 
\qquad+ SW & \hl{13.06} & -1.30 & \hl{10.41} & \hl{7.27} & \hl{24.75} & \hl{12.78} & 1.86 & \hl{11.02} \\ 
SFR-Embed & \hl{\textbf{18.82}} & -0.18 & \hl{5.79} & 2.59 & \hl{24.30} & -5.28 & \hl{7.52} & \hl{18.07} \\ 
\qquad+ SW & \hl{10.46} & \hl{7.33} & \hl{10.91} & \hl{8.12} & \hl{22.50} & \hl{9.62} & \hl{8.22} & \hl{10.47} \\ 
GTE-ModernBERT & \hl{11.64} & \hl{\textbf{12.78}} & \hl{12.98} & 0.74 & 5.07 & \hl{8.82} & \hl{16.20} & \hl{16.77} \\ 
\qquad+ SW & \hl{18.32} & \hl{6.24} & \hl{\textbf{23.59}} & \hl{7.72} & \hl{\textbf{38.21}} & \hl{15.65} & \hl{6.48} & \hl{22.13} \\ 
m2-BERT-32k & \hl{5.26} & \hl{7.81} & \hl{14.55} & 3.75 & -0.11 & -3.62 & \hl{11.10} & \hl{7.80} \\ 
\qquad+ SW & \hl{8.82} & \hl{7.11} & \hl{13.07} & \hl{6.58} & 2.28 & 4.61 & \hl{18.17} & \hl{12.81} \\ 
Voyage-3-large & \hl{14.96} & \hl{7.66} & \hl{16.93} & \hl{11.53} & \hl{21.84} & \hl{\textbf{21.91}} & \hl{15.79} & \hl{\textbf{25.08}} \\ 
\qquad+ SW & \hl{13.99} & \hl{7.04} & \hl{12.51} & \hl{\textbf{15.24}} & \hl{23.54} & \hl{19.04} & \hl{11.61} & \hl{24.41} \\ 
Claude+SW & \hl{6.23} & \hl{8.16} & \hl{7.33} & \hl{13.07} & \hl{21.77} & \hl{5.71} & \hl{9.94} & \hl{8.99} \\
    \bottomrule
  \end{tabular}
  \caption{\label{tab:category-specific}
    When using category-specific instructions, rank-correlation does not show notable improvement and is still quite poor on average. Statistically significant results are  \hl{highlighted}.}
\end{table*}

While capturing author and fandom information is not inherently harmful, the outsized impact of these (trivially computable from metadata) features on embedding-based similarity scores limits their applicability to analysis looking for more subtle phenomena like theme or trope similarity.

\paragraph{Category-specific embeddings show slight improvement.} Table~\ref{tab:category-specific} shows the performance with category-specific instructions across each model and category. When comparing correlations for a given model and context-handling method, category-specific embeddings outperform non-specific embeddings 55\% of the time.

\section{Related Work}
\label{sec:related-work}
\paragraph{Long-context and embedding evaluation.} A number of datasets for long-context evaluation have included literary texts. BookSum \citep{kryscinski-2019} involves summarization over public domain books. LongBench, LongBenchv2, and HELMET \citep {bai-2023, bai-2024, yen2025helmet} include question answering over NarrativeQA \citep{kocisky-etal-2018-narrativeqa}; \citep{zhang2024inftybenchextendinglongcontext} introduces summarization and QA tasks over a set of novels with entity names replaced to reduce the impact of potential contamination. 
Embedding-focused datasets include STS tasks but focus primarily on very short inputs \cite{muennighoff-etal-2023-mteb}; LongEmbed \cite{zhu-etal-2024-longembed}, which evaluates long-context embedding but not on STS, instead using QA tasks over NarrativeQA and SummScreen screenplays \cite{chen-etal-2022-summscreen}. None of these benchmarks measure performance on long-context STS tasks, which are of particular interest to digital humanities and literary scholars \cite{sobchuk-2024}.

\paragraph{NLP tools for literary studies.} A number of works have studied the applicability of NLP methods to digital humanities tasks on public-domain literature.
 \citet{bamman-2024} compare LLMs to traditional supervised methods on a wide range of tasks within literary studies. Other works propose novel computational approaches to analyze elements of fictional texts that are of interest to literary scholars, such as character mobility \citep{wilkens-2024}, emotional arc  \citep{ohman-2022}, and narrative pacing \citep{bamman-2014}. \citet{kohlmeyer-2021} propose \textit{lib2vec}, a method for representing facets of fictional texts using multiple embeddings; because our similarity categories differ, direct application of their method to \textsc{FicSim} is challenging. The (in)applicability of NLP systems to downstream uses has also been studied in other domains, including law \citep{kapoor2024promisespitfallsartificialintelligence} and materials science \citep{gururaja2025textcharacterizingdomainexpert}.
\section{Conclusion}
\label{sec:conclusion}

We present \textit{FicSim}, a dataset of stories and similarity labels for benchmarking model performance on long-context STS tasks within fictional texts. Using \textsc{FicSim}, we show that there is no single model that performs well across all types of similarity---and there are types of similarity for which no model performs well. In corpora with strong superficial similarities, like author or fandom overlap, embeddings may capture this information at the expense of other types of similarity. For this specific type of task, bigger (or more expensive) models are not uniformly better than their smaller, cheaper alternatives. Our evaluation of sliding window attention and category-specific embeddings also demonstrates that sensible modifications to the model to adapt to long-form or literary texts have a limited impact on performance. We call for the careful evaluation of models on the particular task they are applied to, with annotation or validation by subject-matter experts.

The poor performance of otherwise strong models on \textsc{FicSim} highlights the substantial gap that exists between current models and their utility for literary applications. Our models fail to capture finegrained literary similarity and overindex on superficial features of the text in their embeddings. We expect that clever system design or additional domain-specific training could improve performance within this generation of embedding models, and we encourage the evaluation on literary tasks for future embedding model releases.

We hope \textsc{FicSim} will help digital humanities researchers make informed decisions about model selection for tasks relating to story similarity, encourage more evaluation of embedding models on DH tasks, and serve as an example of how creative works can be used for academic research without circumventing creators’ rights and wishes.

\section*{Limitations}

\paragraph{Data collection.}
While we were originally informed by an AO3 support member that leaving comments on fanfics would be an appropriate method for soliciting consent, our account was later temporarily suspended on the basis that we were leaving spam messages. (We had used the same introductory message to reach out to each author, in alignment with IRB protocols.) Our attempt to appeal the suspension was unsuccessful, despite our explanation that we were following instructions we had been given by another AO3 team member. When the temporary suspension was lifted, we decided not to attempt further data collection, because we ultimately did not want to be using AO3 in a way that further increased tension between machine learning researchers and fanfiction writers. 

Thus, while we were able to assemble a dataset using the methods outlined in this project, an exact replication of our process would not be appropriate.

\paragraph{Embedding methods.} It is not possible to consider every possible means of constructing embeddings; while we aimed to capture a set of models and methods that were representative of those applied in digital humanities works with embeddings, it is possible that there exist other methods that would outperform those presented as baselines here. In particular, computing similarities using multiple-embedding strategies is likely to improve performance. We leave devising better embedding strategies for literary domain text to future work. 

\paragraph{Language.} We consider only stories written in English because of our need to reach out to each author individually. 
While we believe the fanfictions within \textsc{FicSim} represent an interesting selection of works across these similarity dimensions, differences exist between literary corpora. It is possible that models that excel at similarity on \textsc{FicSim} would nevertheless struggle on STS tasks for 19th century English literature, short-form satirical poetry from social media, or any other number of specialized literary domains. We see \textsc{FicSim} as an initial step towards improved literary-domain evaluation.

\section*{Acknowledgments}
We extend our deepest gratitude towards the fanfiction authors who have given us permission to include their works in our dataset and without whom this project would not have been possible.

We would also like to thank David Mimno, Maarten Sap, Chap Morack, Suguru Ishizaki, and our reviewers for their helpful feedback on our work.

 AB was supported by a grant from the National Science Foundation Graduate Research Fellowship Program under Grant No. DGE2140739. Any opinions, findings, and conclusions or recommendations expressed in this material are those of the author(s) and do not necessarily reflect the views of the sponsors.

\bibliography{anthology,referencesEMNLP}

\newpage
\appendix

\clearpage

\section{Additional dataset documentation}
\label{appendix:datastats}
\textbf{Tag standardization} Much of our tag categorization and standardization process was inspired by AO3's own practices. AO3 has a team dedicated to \textit{tag wrangling}, which is the task of maintaining a database of canonical tags, sorting and organizing those tags, and linking tags to their canonicalized form. Thanks to this high standard of organization, many tags can be mapped back to canonical tags. Non-canonical tags often come in the forms of meta commentary, merging of multiple canonical tags, or a modification of canonical tags to include fandom-specific references. These tags still contain valuable information about their stories, and looking at them in conjunction with similar canonical tags sometimes helped us determine appropriate categorizations.

\textbf{Comparisons by category.} Not all stories have a similarity score along every axis. Table~\ref{tab:num_comparisons} lists the number of comparisons possible in each category. 

\begin{figure}
    \centering
    \includegraphics[width=1.0\linewidth]{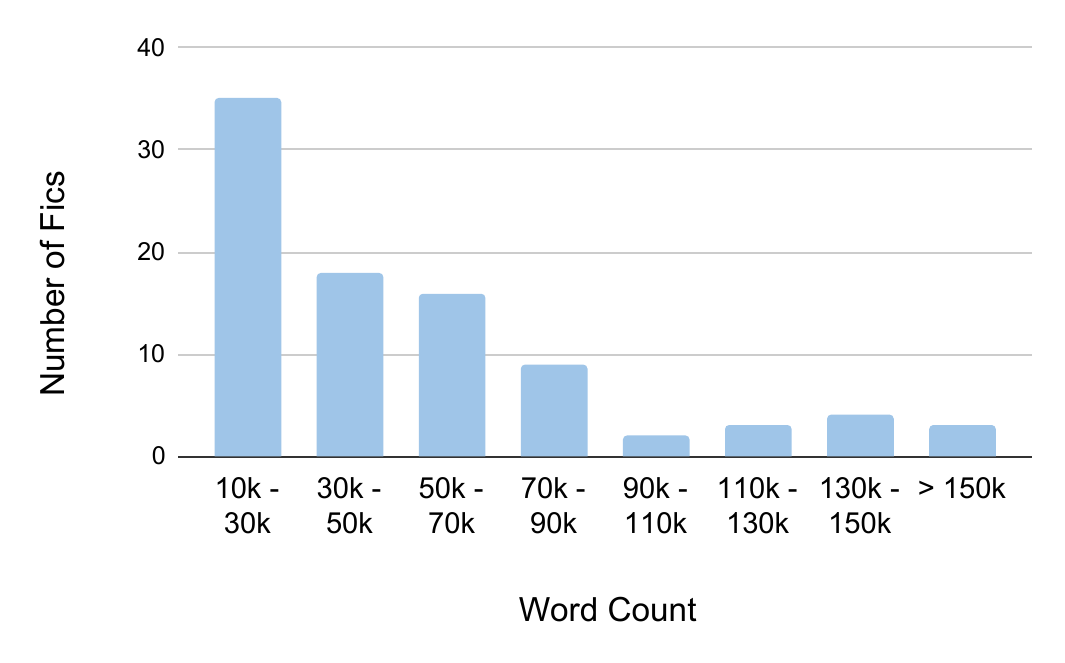}
    \caption{Story lengths in FicSim range from 10k to >400k words.}
    \label{fig:input-lengths}
\end{figure}

\begin{table}
  \centering
  \small
  \begin{tabular}{lr}
    \toprule
    \textbf{Comparison axis} & \textbf{Pair count} \\
    \midrule
    Plot & 4005\\
    Character States & 3240 \\
    Relationship Dynamics & 2278 \\
    Theme & 1431\\
    Time & 105 \\
    Style & 1431\\ 
    Fanfiction Tone \& Content Tags & 1275 \\
    Overall (All fandom-agnostic tags) & 4005\\
    Overall (not fandom-agnostic tags) & 4005\\
    Fandom-Specific Tags & 4005 \\
    Fandoms & 4005 \\
    Author & 4005 \\
    
    \midrule
    Total  & 33,790\\
    
    \bottomrule
  \end{tabular}
  \caption{\label{tab:num_comparisons}  The number of pairwise comparisons in \textsc{FicSim} by category. We exclude stories from pairwise comparisons in categories where they lack tagging. 
  }
\end{table} 
\textbf{Length}
Figure~\ref{fig:input-lengths} shows the length distribution of texts in \textsc{FicSim}. 

\textbf{Additional metadata.} In addition to the similarity scores and full texts, \textsc{FicSim} contains many other metadata fields about each story, including chapter splits, author IDs, author-written summaries (where available), and a number of AO3-imposed classifications (e.g. the genders of the characters in the primary relationship in the story). While we do not explicitly clean data for additional non-tag categories, we make these data available in the hope that they will be useful to other researchers working on literary applications. 
\clearpage
\section{Additional documentation of models}

\label{appendix:models}
\begin{table*}
  \centering
  \small
  \begin{tabular}{llllll}
    \toprule
    \textbf{\makecell[l]{Abbreviated name}} & \textbf{HF or API name} & \textbf{\makecell[l]{Max\\ context}} & \textbf{Pooling strategy} & \textbf{Param count} \\
    \midrule
    Linq-Embed & Linq-Embed-Mistral & 32,768 & Last-token &   7B\\
    SFR-Embedding & SFR-Embedding-Mistral & 32,768 & Last-token &  7B \\
    GTE-Qwen2 & gte-Qwen2-7B-instruct & 32,768 & All tokens&  7B \\
    GTE-ModernBERT & gte-modernbert-base  & 8,192 & CLS & 149M & \\ \midrule
    m2-BERT-32k & m2-bert-80M-32k-retrieval &   32,768 & CLS   &  80M \\
    Voyage-3-large & voyage-3-large & 32,768 & Unknown & Unknown \\
    Claude-3.7-Sonnet & claude-3-7-sonnet-20250219 & 200,000 & n/a & Unknown \\
    
    \bottomrule
  \end{tabular}
  \caption{\label{tab:model-details}
    Additional details on the models ran. 
  }
\end{table*} 
This section contains additional details for reproducing the embedding methods.

\paragraph{Models.} We evaluate on 7 models, described in detail in Table~\ref{tab:model-details}.

\textbf{Pooling.} We follow each model's default strategy for constructing embeddings: Linq-Embed and SFR-Embedding use last-token pooling, GTE-Qwen2 uses mean pooling of all tokens, and GTE-ModernBERT uses a CLS token. In cases where we obtain multiple embeddings (i.e., when using a sliding window), we average all embeddings to produce a single embedding for each document. When mean-pooling multiple embeddings, we average \textit{after} pooling all token embeddings instead of averaging a single pooled embedding from each window. In mean-pooled embeddings, we include prompt tokens but take only one embedding of each overlapped token in sliding windows with overlap.

\paragraph{Software.} For single-window (truncation) approaches, we use \texttt{sentence\_transformers} \cite{reimers-gurevych-2019-sentence}. For sliding window approaches, we use Hugging Face \texttt{transformers} \cite{wolf2020huggingfacestransformersstateoftheartnatural}. We call Voyage and Claude through their respective APIs, which (at the time of writing) do not retain user data for model training. A limited amount of language model assistance was used for writing simple data processing scripts; all code was verified by the authors.

\paragraph{Computational resources.} All local models were run on a mixture of L40S and H100 GPUs; we estimate that the total compute time in development and running the final embedding methods did not exceed 200 GPU-hours. The total cost of development and running the API-based methods was approximately \$80, of which \$78 was the cost of running Claude. 

\paragraph{Prompts.} We use the same prompt for all models except Claude; the category-specific prompts are in Table~\ref{tab:embedding-prompts}. For Claude, we use the system prompt ``Below is a long-form fanfiction written in English. You will be asked to summarize this story.'' We provide the full text of the story as a user message, then provide an additional user message with instructions. The instruction message always begins ``Please write a detailed summary of this story, using up to 5,000 words.'' It optionally also has a category-specific instruction; these instructions are listed in Table~\ref{tab:claude-prompts}. 

In the rare (3) cases where a story exceeds Claude's context window, we summarize as much of the story as possible in a first API call and provide the summary plus the remainder of the story in a second API call. In this second call, the system prompt is changed to ``Below is a long-form fanfiction written in English. The first section is a summary of the first portion of the story, and then the remainder of the story follows. You will be asked to summarize this story.'' The two sections of the input are labeled ``Summary:'' and ``Remainder of the story:'' and separated by a line of dashes. The instruction is changed to ``Please write a detailed summary of the full story, using up to 5,000 words. You may copy the summary of the first portion of the story exactly, or modify it as you wish.'' along with any category-specific instruction. 

\begin{table*}
    \begin{tabular}{l|p{0.7\textwidth}}
         \toprule
         \textbf{Category} &  \textbf{Prompt} \\ \midrule
        Plot &   Identify the main plot arc of the fanfiction based on the text.
\\
        Character state &     Identify the main character states of the fanfiction based on the text.
\\
        Relationship dynamic &     Identify the main relationship dynamics of the fanfiction based on the text.
\\ 
        Theme &     Identify the main themes of the fanfiction based on the text.
\\ 
        Time &     Identify the main time period of the fanfiction based on the text.
\\ 
        Style &     Identify the main literary style of the fanfiction based on the text.
\\ 
        Tone \& content &     Identify the main fanfiction-specific tone and content descriptors of the fanfiction based on the text.
\\
        Overall & [no prompt] \\
         \bottomrule
    \end{tabular}
    \caption{Prompt for all embedding models. The prompt (with any applicable model-specific formatting) is prepended to the beginning of the text and the start of every sliding window.}
    \label{tab:embedding-prompts}
\end{table*}

\begin{table*}
    \begin{tabular}{l|p{0.7\textwidth}}
         \toprule
         \textbf{Category} &  \textbf{Prompt} \\ \midrule
        Plot &   In your summary, pay particular attention to the plot of the text.
\\
        Character state &    In your summary, pay particular attention to the attributes of the characters in the text.
\\
        Relationship dynamic &     In your summary, pay particular attention to the relationship dynamics of the characters in the text.
\\ 
        Theme &     In your summary, pay particular attention to the themes of the text.
\\ 
        Time &    In your summary, pay particular attention to the temporal setting of the text.
\\ 
        Style &     In your summary, pay particular attention to the literary style of the text.
\\ 
        Tone \& content &     In your summary, pay particular attention to any fanfiction-specific tone or tropes exhibited in the text.
\\
        Overall & [no additional prompt] \\
         \bottomrule
    \end{tabular}
    \caption{Prompt for Claude summarization. This is appended as part of the last user message, after the system message and a user message containing the full text of the story.}
    \label{tab:claude-prompts}
\end{table*}

\clearpage
\section{Author Consent Process}
\label{appendix:author-consent-forms}

Archive of Our Own does not have a private messaging feature, and authors do not generally post contact information (or real names) on their fanfictions. After consulting with the AO3 policy team, we agreed to reach out to authors by leaving a comment on the stories we would like to use. This comment then directs them to our main Reddit post, which links to the project's webpage, explains the study and its terms, and offers a locale for authors to ask questions directly of the authors. This process was approved as Carnegie Mellon University IRB Study 00000260.

\textbf{Revoking consent.} We maintain a Google Form for requesting removal of a story at any time, with no questions asked. We commit to monitoring this form in perpetuity and removing fanfiction promptly if authors choose to revoke consent. For replication of results on the dataset, we will clearly label the dataset on Hugging Face and the repository with a version number, and ask that anyone using the dataset report the evaluation version.

\subsection{Outreach process documents}
We provide the exact text of the comments to reach out to authors (Figure~\ref{fig:comment-to-authors}) and the text of the Reddit post (Figure~\ref{fig:redditpost}). \footnote{Our original post cites 30k as our desired lower word limit for fanfiction contributions, but after seeing the volume and quality of fanfic contributions we received below this threshold, we decided to include texts above 10k words in our dataset, provided they had adequately detailed tagging.}

\begin{figure*}[h]
\begin{tcolorbox}[colback=gray!10!white, colframe=gray!50!black, boxrule=0.5pt, sharp corners]

Hi! My name is Natasha Johnson :) I’m a recent graduate from Carnegie Mellon University’s English Department. I’m working alongside Emma Strubell and Amanda Bertsch at CMU on a project involving fanfiction
, and we’re hoping to include your fanfic(s) in our research. If you would like to learn more about our project and consent for us to include your work, please take a look at the post we made about it here:  \url{https://www.reddit.com/r/AO3/comments/1fu8m6u/fanfiction_study_recruitment/}
\end{tcolorbox}
\caption{Sample comment on fanfiction}
\label{fig:comment-to-authors}

\end{figure*}

\begin{figure*}
\begin{tcolorbox}[colback=gray!10!white, colframe=gray!50!black, boxrule=0.5pt, sharp corners]

Hello! We are Natasha Johnson (\url{https://natashamariejohnson330.github.io/}), Emma Strubell (\url{https://strubell.github.io/}), and Amanda Bertsch (\url{https://www.cs.cmu.edu/~abertsch/}). 
We're interested in exploring the capabilities and limitations of digital tools in the context of humanities research. We are currently conducting a research project that looks at quantifying fanfiction similarity, focusing on fics over 30k words.

Because of the detailed tagging you use on your work, we're asking for your consent to use your fanfiction for this project.

If you consent to us using your fanfic(s), here are our promises:
\begin{enumerate}
    \item We might make observations about fanfic content, but we will not critique fanfics in any way.
    \item We will actively avoid seeking any personal information about you.
    \item We will not use your fanfics to train AI models.
    \item We will use your fanfics to test how well AI models capture similarity in literary contexts, to see if these models could be useful for literary scholars. During testing, the models do not retain any history or memory of input text, and the models are not trained on the inputs.
    \item If we publish our research, we will release our dataset alongside it. This dataset will include the fanfic texts, the fanfiction tags, and a numerical author identifier in place of your AO3 pseudonym. 
    \item In order to access the dataset, we will ask viewers to agree not to use the data for AI training purposes.
    \item At your request, we will remove your fanfic(s) from the dataset at any time, for any reason. Here’s the form you can use to submit a removal request: \url{https://forms.gle/JxHRsTbwMb78fCS77}
\end{enumerate}

If you would like to give us permission to use your fanfic(s) in this way, please let us know via this consent form: \url{https://forms.gle/GrU86ZWEWm1kvY2C6}

Feel free to post any questions here in the comments, or you can reach out anonymously via this Google form:  \url{https://forms.gle/94ejTXtwxZyVriqh8}
\end{tcolorbox}
\caption{Post on Reddit with information on how to consent}
\label{fig:redditpost}

\end{figure*}

\clearpage
\section{License}
\label{appendix:license}

Copyright 2025, the original author of each fanfiction (used with permission).

Permission is hereby granted, free of charge, to any person obtaining a copy of this software and associated documentation files (the “Software”), to deal in the Software without restriction, including without limitation the rights to use, copy, modify, merge, publish, distribute, sublicense, and/or sell copies of the Software, and to permit persons to whom the Software is furnished to do so, subject to the following conditions:

No part of the text of any story in the dataset will be used in training of any machine learning model, or in any system that involves a model retaining memory, knowledge, or other influence from the story text. 

The above copyright notice and this permission notice shall be included in all copies or substantial portions of the Software.

THE SOFTWARE IS PROVIDED “AS IS”, WITHOUT WARRANTY OF ANY KIND, EXPRESS OR IMPLIED, INCLUDING BUT NOT LIMITED TO THE WARRANTIES OF MERCHANTABILITY, FITNESS FOR A PARTICULAR PURPOSE AND NONINFRINGEMENT. IN NO EVENT SHALL THE AUTHORS OR COPYRIGHT HOLDERS BE LIABLE FOR ANY CLAIM, DAMAGES OR OTHER LIABILITY, WHETHER IN AN ACTION OF CONTRACT, TORT OR OTHERWISE, ARISING FROM, OUT OF OR IN CONNECTION WITH THE SOFTWARE OR THE USE OR OTHER DEALINGS IN THE SOFTWARE.

\end{document}